\pdfoutput=1

\documentclass[11pt]{article}

\usepackage[final]{acl}

\usepackage{times}
\usepackage{latexsym}

\usepackage[T1]{fontenc}

\usepackage[utf8]{inputenc}

\usepackage{microtype}

\usepackage{inconsolata}

\usepackage{graphicx}

\usepackage{amsmath}
\usepackage{amssymb}
\usepackage{mathtools}
\usepackage{amsthm}

\usepackage{algorithm}
\usepackage{algorithmic}
\usepackage{hyperref}
\usepackage[capitalize,noabbrev]{cleveref}
\usepackage{booktabs}
\usepackage{subcaption}
\usepackage{xspace}
\usepackage{multirow}
\usepackage{xcolor}

\title{Efficiently Editing Mixture-of-Experts Models with Compressed Experts}

\author{Yifei He$^{1}$\thanks{Work done during an internship at Microsoft.} \quad Yang Liu$^{2}$ \quad Chen Liang$^{2}$ \quad Hany Hassan Awadalla$^{2}$\\
  $^{1}$University of Illinois Urbana-Champaign 
  \quad $^{2}$ Microsoft \\
  \texttt{yifeihe3@illinois.edu} \\
  \texttt{\{yaliu10,chenliang1,hanyh\}@microsoft.com}
}

\begin{document}
\maketitle
\begin{abstract}
Mixture-of-Experts (MoE) models have become a key approach for scaling large language models efficiently by activating only a subset of experts during training and inference. Typically, the number of activated experts presents a trade-off: fewer experts reduce computational costs, while more experts improve performance. Recent studies reveal that not all activated experts contribute equally to model performance, with some providing minimal utility, particularly when finetuning pretrained MoE models for specialized downstream tasks. The co-existence of significant and redundant  parameters in experts provides us an opportunity to reduce the number of activated experts while maintaining model performance. In this work, we propose the concept of \textit{compressed experts}, lightweight modules that serve as compact representations of full experts. Our approach preserves the most important experts while replacing other auxiliary activated experts with compressed experts. The reduction of active parameters significantly lowers inference costs while achieving comparable performance. Extensive experiments on models including Phi-MoE and OLMoE demonstrate that compressed experts recover over 90\% of full expert performance across various tasks while reducing more than 30\% active parameters and saving 20\% in inference costs. This approach enables efficient deployment of MoE models in resource-constrained settings and facilitates scaling to larger models with manageable overhead. Our code is available at \url{https://github.com/yifei-he/Compressed-Experts}. \looseness=-1
\end{abstract}

\section{Introduction}
Mixture-of-Experts (MoE) models have emerged as an effective approach to scale up the sizes of large language models (LLM) with minimal computational overhead~\citep{shazeer2016outrageously,lepikhingshard,fedus2022switch,cai2024survey}. In Transformer-based MoE models, the feed-forward networks (FFNs) are replaced by MoE layers, each containing multiple experts. For a given input, a routing network routes it only to a selected subset of relevant experts, ensuring that only a fraction of the network is activated during each forward pass. This sparse activation significantly reduces computational costs compared to dense models while maintaining high model capacity.

\begin{figure}[t!]
    \centering
    \includegraphics[width=0.8\linewidth]{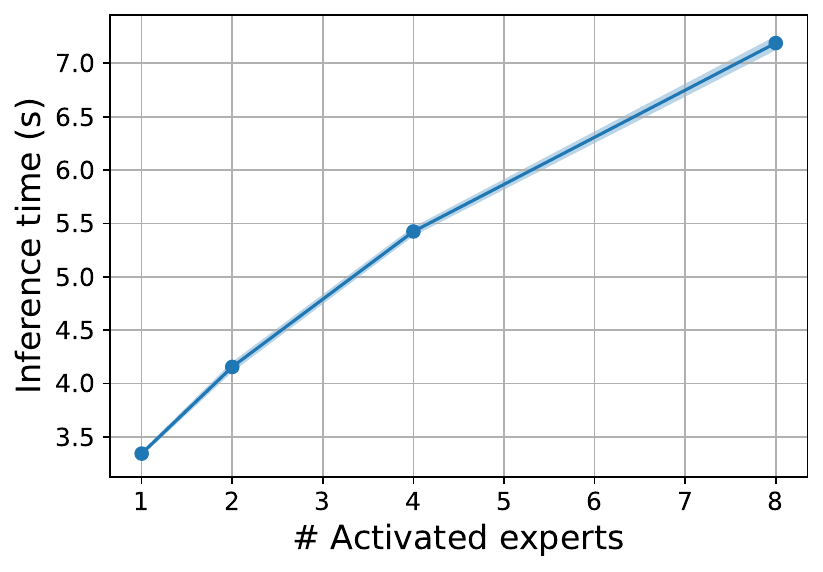}
    \caption{The inference time of MoE models grows linearly with increasing number of activated experts. The plot is generated using OLMoE architecture. Details on the inference speed measurement methodology are provided in \Cref{sec:setup}.}
    \label{fig:linear_inference}
    \vspace{-0.5cm}
\end{figure}

In practice, MoE models often require multiple activated experts to achieve desirable performance. For instance, Mixtral~\citep{jiang2024mixtral} and Phi-MoE~\citep{abdin2024phi} activate two experts, Qwen-MoE~\citep{yang2024qwen2technicalreport} activates four and OLMoE~\citep{muennighoff2024olmoe} activates eight. The efficiency of MoE models largely depends on the number of activated experts. Specifically, the inference cost of MoE models increases linearly with this number (as shown in \Cref{fig:linear_inference}). Fewer activated experts reduce training and deployment costs as a smaller portion of the network is used. Conversely, activating more experts increases utilization of model capacity, often leading to superior performance. Thus, the selection of the number of activated experts poses a fundamental trade-off between efficiency and performance. \looseness=-1

Recently, several studies highlight the potential \textit{redundancy} in activated experts. For instance, \citet{lepikhingshard} demonstrates that increasing the number of activated experts yields diminishing returns in performance gains. Additionally, \citet{huang2024harder} shows that not all tasks necessitate full utilization of all the top-$k$ experts. These findings suggest that only a subset of the activated experts, which we term \textit{main experts}, contribute significantly to model utility, while others act as \textit{auxiliary experts} with limited impact on performance. Despite their lower contribution, passing through auxiliary experts incurs the same computational cost as main experts. This redundancy is particularly pronounced when finetuning a pretrained MoE model on specific downstream tasks, where expert utility varies depending on task difficulty~\citep{huang2024harder}. This inefficiency motivates our approach: compressing auxiliary experts to reduce computational overhead while maintaining model performance. To achieve this, we outline three design principles: \textbf{i) Efficiency}: The module should be computationally efficient, ensuring minimal overhead for training and inference. \textbf{ii) Expressiveness:} It should retain the capabilities of the auxiliary experts to minimize performance loss. \textbf{iii) Flexibility:} The module should adapt to various expert combinations, as the top-$k$ experts may differ across tokens within the same sequence. 

\begin{figure}[t!]
    \centering
    \includegraphics[width=0.9\linewidth]{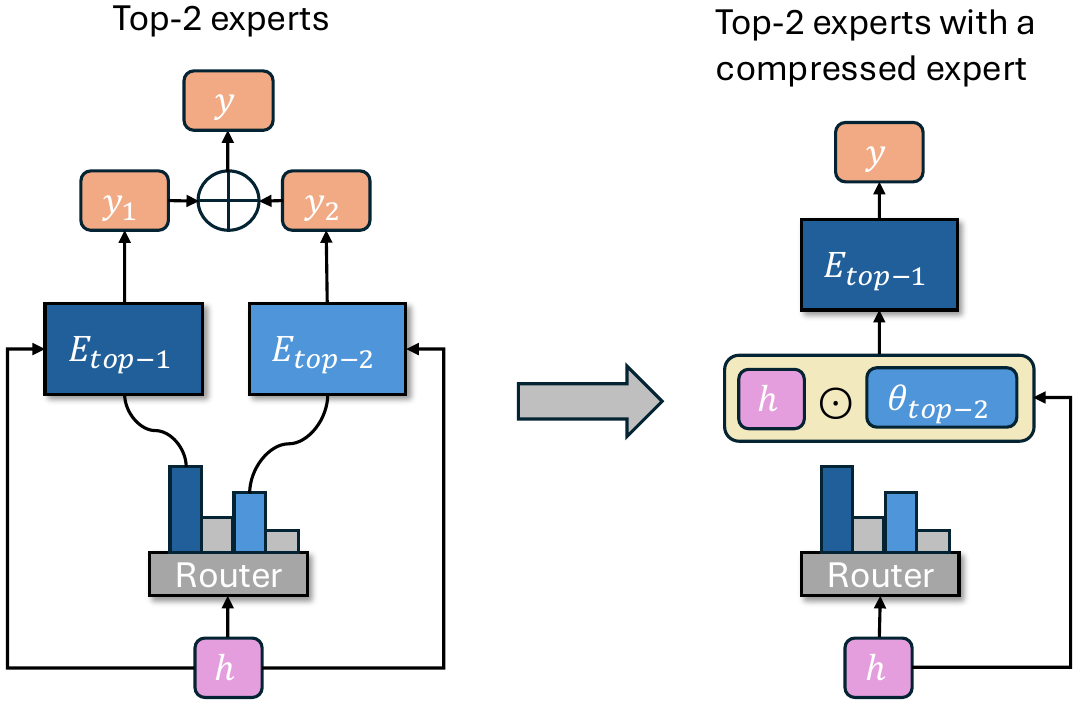}
    \caption{Reducing a top-2 MoE layer to top-1 with compressed experts. Our approach replaces the second expert $E_{\text{top-2}}$ with a compressed expert $\theta_{\text{top-2}}$, and augment the hidden state $h$ via element-wise multiplication. This enables a single forward pass through one expert instead of two, significantly reducing inference cost while maintaining comparable performance.}
    \label{fig:moe}
    \vspace{-0.5cm}
\end{figure}

Following these principles, we propose \textit{compressed experts}, which are embedding vectors serving as compact representations of auxiliary experts. In MoE layers, we introduce one compressed expert per full expert, maintaining a one-to-one correspondence. These compressed experts are extremely lightweight, with a dimensionality matching the hidden states and consisting of less than 0.05\%\footnote{This ratio may vary for different MoE models as it depends on the specific configurations of FFNs. Here, 0.05\% is computed based on the configurations of Phi-MoE.} of the parameters of a full expert. During forward passes, the compressed experts corresponding to auxiliary experts are aggregated via a weighted sum and incorporated into the hidden states through element-wise multiplication. Instead of activating both main and auxiliary experts, the augmented hidden states are processed only through main experts, significantly reducing computation while preserving contributions of auxiliary experts. We illustrate this approach in Fig. \ref{fig:moe}, with a top-2 MoE layer containing one main expert and one auxiliary expert. In this example, the second expert is replaced by a compressed expert, reducing the number of forward passes from two to one.


Through extensive experiments on popular MoE models including Phi-MoE~\citep{abdin2024phi} and OLMoE~\citep{muennighoff2024olmoe}, we demonstrate that compressed experts recover over 90\% of full-expert performance across diverse tasks while reducing more than 30\% of active parameters and cutting inference cost by 20\%. By bridging the gap between performance and efficiency, compressed experts enable scalable and cost-effective deployment of MoE models, making them more practical for resource-constrained environments.

\section{Method}

\subsection{MoE Layers} \label{sec:background}
Transformer-based Mixture-of-Experts (MoE) models extend standard Transformer architectures by replacing feed-forward network (FFN) layers with MoE layers, each comprising a set of $n$ experts. For a given input sequence, the hidden state $h$ of each token is processed by a routing network $R$, which computes a routing weight $\alpha_i$ with respect to each expert $E_i$. Here, we focus on top-$k$ routing~\citep{shazeer2016outrageously}, which selects $k$ experts for each token to be active in forwarding. The MoE layer output is then computed as a weighted sum of the outputs from the $k$ experts selected based on their routing weights:
\begin{align} \label{eq:moe_ffn}
    y=\sum_{i=1}^{n} \alpha_i\cdot E_i(h),
\end{align}
where $\alpha_i=\text{top-}k(\text{Softmax}(R(h)))_i$. In this formulation, the top-$k$ mechanism ensures sparse activation, where only the $k$ experts with the highest routing weights contribute to the output, while the remaining $n-k$ experts are effectively disabled ($\alpha_i=0$). Typically, only a small fraction of experts is utilized during each forward pass, i.e., $k\ll n$, enabling substantial computational savings. For instance, Phi-MoE~\citep{abdin2024phi} activates 2 out of 16 experts, and OLMoE~\citep{muennighoff2024olmoe} activates 8 out of 64 experts, etc. This sparse activation significantly reduces computational costs compared to dense models, making MoE layers efficient and scalable. 

Despite these efficiency gains, the predetermined number of activated experts $k$ might introduce inherent redundancy. Eq.~\ref{eq:moe_ffn} requires activating all $k$ selected experts, but some of which may contribute minimally or partially to model predictions. This has been especially observed on specialized tasks, where only some main experts dominate the routing weights while auxiliary experts have very low weights~\citep{huang2024harder}. An alternative is to reduce the number of $k$ for customized scenarios after model pretraining. Compressed Experts, a lightweight module that provides vectorized expert features to main experts, can effectively edit a MoE model with a reduced number of expert computations while preserving  performance.


\begin{figure}[t!]
    \centering
    \includegraphics[width=\linewidth]{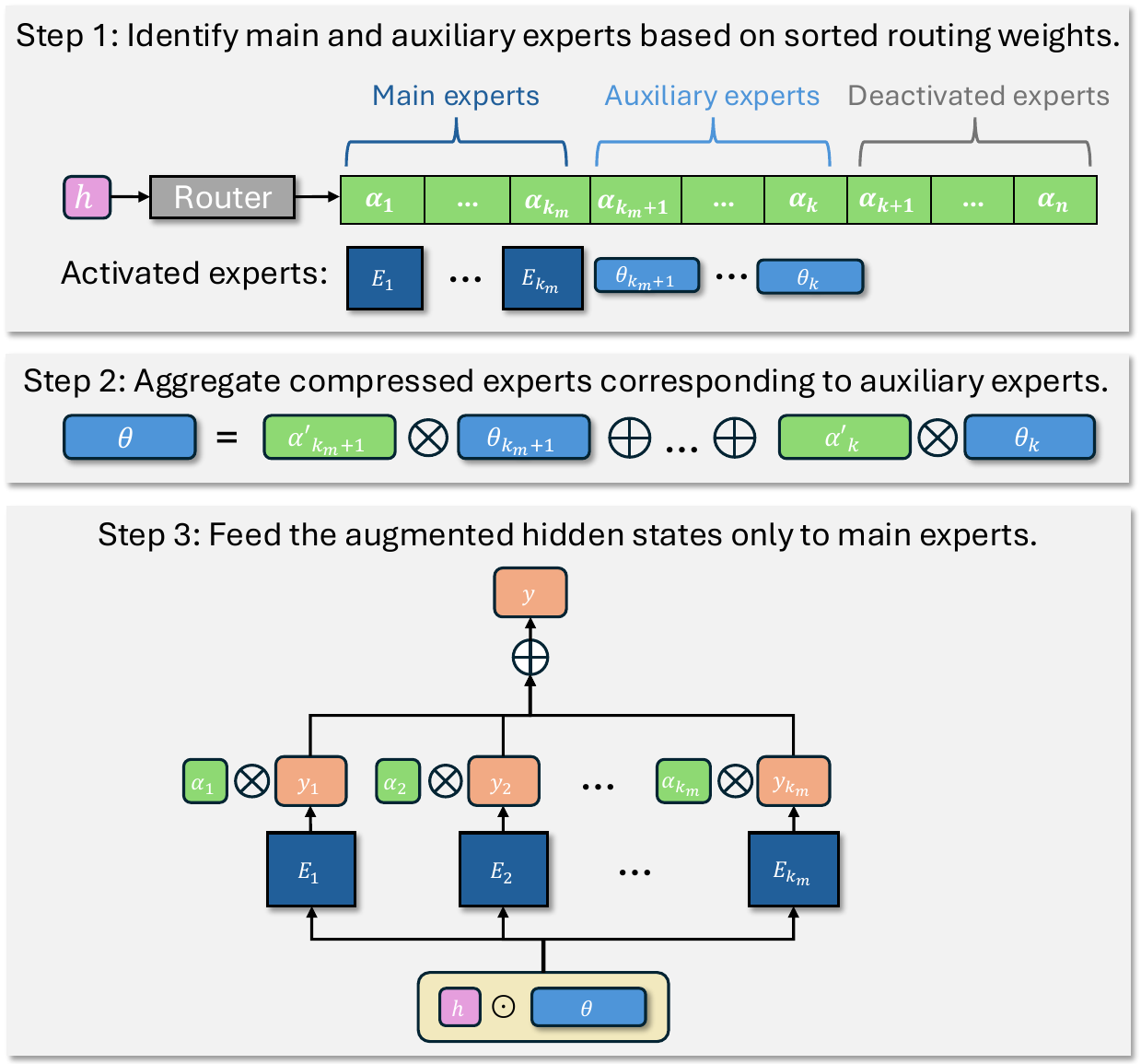}
    \caption{The compressed expert integration process. \textbf{Step 1}: The router selects the top-$k$ experts, categorizing them into $k_m$ main experts and $k_a$ auxiliary experts based on routing weights. \textbf{Step 2}: The compressed representations of auxiliary experts $\{\theta_i\}_{i=k_m+1}^k$ are aggregated through a weighted sum using normalized routing weights $\{\alpha_i'\}_{i=k_m+1}^k$. \textbf{Step 3}: The hidden state $h$ is augmented with the aggregated compressed expert $\theta$ and passed only through main experts, reducing computation while maintaining performance.}
    \label{fig:moe_detail}
    \vspace{-0.3cm}
\end{figure}

\subsection{Compressed Experts} \label{sec:ce}
Each MoE layer consists of $n$ total experts, each of which is a feed-forward network (FFN), denoted as $\{E_i\}_{i=1}^n$. To enhance computational efficiency, we introduce the same number of compressed experts $\{\theta_i\}_{i=1}^n$, each serving as a compact representation of its corresponding full expert. The compressed experts are \textit{embedding vectors} with the same dimension as the hidden states, i.e., $\theta_i\in\mathbb{R}^d$. 

Conventionally, for each input token, $k$ out of the $n$ experts are activated. We can categorize these activated experts into  $k_m$ main experts and $k_a=k-k_m$ auxiliary experts, based on their ordered routing weights. In many scenarios, instead of fully activating all $k$ experts, only forwarding the main experts can lead to a better performance-cost trade-off. However, we do not want to completely discard the information in auxiliary experts.
Compressed experts provide a solution to leverage their compressed counterparts to reduce computational cost while retaining their contributions. We introduce it with the following three steps (illustrated in \Cref{fig:moe_detail}).\looseness=-1

\textbf{Expert identification}: For an MoE layer, given an input hidden state $h$, the router outputs $n$ routing weights with respect to each expert: $\{\alpha_i\}_{i=1}^n$. The top $k_m$ experts with the highest routing weights are designated as main experts, while the next $k_a$ are identified as auxiliary experts. The remaining experts are not activated for this input.

\textbf{Compressed experts aggregation}: Instead of directly using auxiliary experts, we approximate their contribution using their compressed embedding vectors. For efficient computation, we aim to combine the information contained in each of the compressed expert, and make it compatible with element-wise computation with the hidden states. To achieve this, we aggregate the compressed experts through a weighted sum, where each auxiliary expert's compressed representation is scaled by its normalized routing weight: $\theta = \sum_{i={k_m+1}}^k \alpha_i' \cdot \theta_i$. Here, $\alpha_i'$ represents the normalized routing weights such that $\sum_{i={k_m+1}}^k \alpha_i'=1$, ensuring that the aggregated compressed expert retains the overall weight distribution. \looseness=-1

\textbf{Hidden states augmentation}: The aggregated compressed expert $\theta$ is incorporated into the hidden state $h$ through an element-wise product. This operation has proven highly effective by PEFT methods such as (IA)$^3$~\citep{liu2022few}, which applies a similar transformation to attention activations. Then, the augmented hidden state is passed only through the main experts, eliminating the need for forward computation through the $k_a$ auxiliary experts, thereby reducing inference cost. Note that using an alternative approach that modifies model parameters dynamically would introduce substantial complexity, as it would require constructing a new model configuration for each token in a sequence on the fly, given the combinatorial number of possible expert selections (i.e., choosing k active experts from n total experts). The element-wise product avoids this issue by enabling seamless integration of compressed experts without disrupting token-level independence.

For stability around initialization, all $\theta_i$ values are initialized as ones. Combined with the normalized routing weights in step 2, this ensures that $\theta$ remains identity-like at initialization, preventing drastic changes to hidden state transformations during early training.

\subsection{Expert Reduction} \label{sec:km_ka}
While compressed experts are flexible and can be applied to any number of main and auxiliary experts, we investigate the conditions under which they are most effective. Specifically, we use OLMoE, which has 8 active experts ($k=8$), as a case study. We evaluate different configurations by varying the number of main experts ($k_m\in\{1,2,4,8\}$), and replace the remaining $k_a=k-k_m$ auxiliary experts with compressed experts (abbreviated as CE). To assess the impact of compressed experts in downstream adaptation, we conduct experiments using supervised finetuning (SFT) on pretrained OLMoE with mathematical data, followed by evaluation on GSM8k with 0-shot CoT (details provided in \Cref{sec:setup}).

\begin{table}[t!]
\centering
    \scalebox{0.9}{\begin{tabular}{ccc}
        \toprule
        \textbf{Configuration} & \textbf{w/o CE }& \textbf{w/ CE}\\
        \midrule
        Top-1 & 12.4 & 15.2 \\
        Top-2 & 18.0 & 20.6\\
        Top-4 & 26.8 & 29.7\\
        Top-8 & 32.3 & /\\
        \bottomrule
    \end{tabular}}
    \caption{GSM8K 0-shot CoT exact match scores (\%) for OLMoE with varying numbers of main experts. Compressed experts (CE) improve performance across reduced configurations, but cannot fully recover the performance lost when reducing more than half of the experts.}
    \label{tab:n_experts}
    \vspace{-0.3cm}
\end{table}

As shown in \Cref{tab:n_experts}, compressed experts consistently improve performance across reduced configurations, demonstrating their ability to recover performance lost due to expert reduction. However, their effectiveness is limited when more than half of the experts are treated as auxiliary.

For instance, in the top-2 setting, two experts function as main experts, while the remaining six are treated as auxiliary experts. Using compressed experts improves performance in this case, but top-2 with CE (20.6\%) noticeably lags behind top-4 without CE (26.8\%). This indicates that the top-4 experts likely all contribute significantly to model performance, and compressing too many main experts leads to significant performance degradation. 
This highlights the importance of main and auxiliary expert categorization to maximize efficiency without sacrificing performance.

This result suggests that compressed experts are most effective when halving the number of activated experts. Also empirically, in our experiments, structuring the model such that main and auxiliary experts are evenly split ($k_m=k_a=k/2$) strikes a good balance between efficiency and performance, retaining 92\% of the full configuration’s performance while reducing computational cost. Further reductions lead to noticeable performance degradation, as critical expert contributions become increasingly difficult to compress effectively. 


\subsection{Reduction in Activate Parameters} \label{sec:active_params}

Replacing half of the activated experts with compressed experts significantly reduces the number of active parameters. Taking Phi-MoE as an example, each MoE layer consists of three weight matrices of size $4096\times 6400$. The total non-MoE parameters are approximately 2.4B. In the original configuration with 2 activated experts across 32 MoE layers, the total number of MoE parameters is $3\times4096\times6400\times32\times2\approx5.03B.$
With compressed experts, only 1 expert is activated per layer, and the compressed experts add minimal parameters (underlined below): $3\times(4096\times6400\times1+\underline{2\times4096+6400})\times32\approx2.52B$. In total, in one forward pass, the number of active parameters for the original configuration is 7.45B, while the one with compressed experts is only 4.93B, resulting in a saving of 33.8\%. We provide a similar calculation for OLMoE in \Cref{appendix:params}, which results in a 31.4\% reduction in active parameters.

\begin{figure}[t!]
    \centering
    \includegraphics[width=0.7\linewidth]{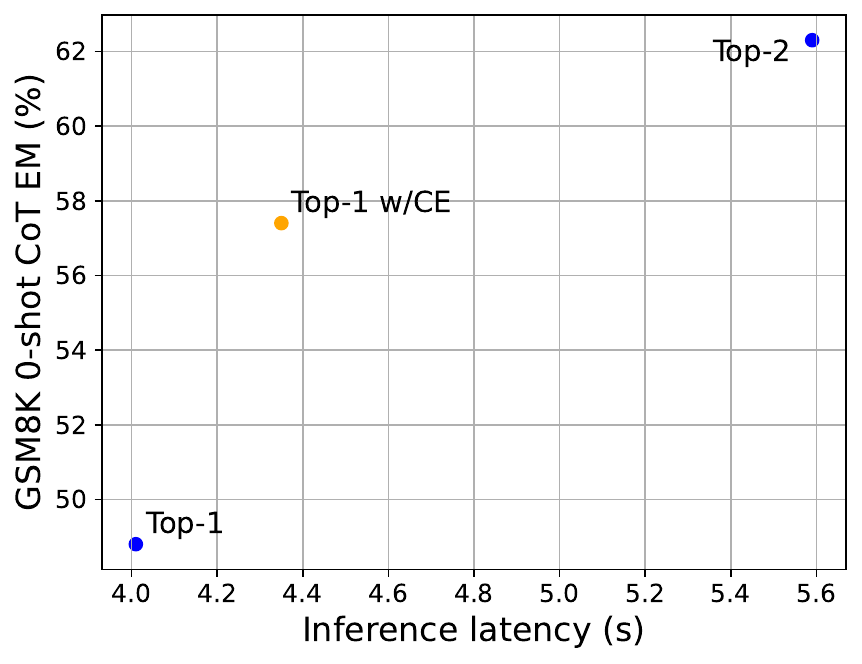}
    \caption{The performance of Phi-MoE versus the inference latency, each point representing a different expert configuration. The Top-1 w/ CE configuration performs closely to the Top-2 configuration while achieving low inference latency close to Top-1. A similar plot for OLMoE is in \Cref{appendix:perf_time}.}
    \label{fig:perf_time}
    \vspace{-0.3cm}
\end{figure}

\subsection{Performance-Latency Trade-off} \label{sec:emp_val}
To analyze the effectiveness of compressed experts, we compare the performance and inference latency of three configurations: top-2, top-1 with compressed experts (abbreviated as CE), and top-1, as shown in \Cref{fig:perf_time}. The performance is evaluated on the Phi-MoE finetuned on mathematical reasoning data, with detailed training and evaluation procedures described in \Cref{sec:setup}. 

The results illustrate a clear trade-off between performance and inference latency across the configurations. The top-2 configuration achieves the highest performance  with a score of 62.3\%, but this comes at the cost of the highest inference latency of 5.59 seconds. In contrast, the top-1 configuration is the most computationally efficient, with a latency of 4.01 seconds, but its performance is significantly lower at 48.8\%. This gap between the two configurations highlights the trade-off between computational efficiency and model capacity utilization. \looseness=-1

Introducing compressed experts in the top-1 w/ CE configuration effectively addresses this trade-off. With a latency of 4.35 seconds, the compressed expert configuration adds only a minimal computational overhead compared to top-1. However, it achieves a substantial performance improvement over top-1, closing much of the gap between top-1 and top-2 by reaching a performance of 57.4\%. This demonstrates the ability of compressed experts to augment the utility of the top-1 expert with minimal additional cost, making it a practical and efficient model editing approach.

\section{Experiments}
To comprehensively evaluate the effectiveness of compressed experts, we integrate them into MoE models during the supervised finetuning (SFT) stage. We focus on the common scenario where a model is pretrained with a fixed number of activated experts and later adapted for downstream tasks. This setting is particularly relevant for compressed experts, as the redundancy of auxiliary experts becomes more pronounced during finetuning, when only a subset of experts may be crucial for the specific task.

In \Cref{sec:setup}, we detail the experimental setup, including model configurations, datasets and training specifics. In \Cref{sec:results}, we first present the performance of compressed experts on evaluation benchmarks, then show their the inference cost savings, demonstrating that compressed experts achieves an effective balance between performance and efficiency. Finally, in \Cref{sec:analysis}, we provide a detailed analysis on the task-specific performance of compressed experts.

\subsection{Setup} \label{sec:setup}

\begin{table}[t]
    \centering
    \scalebox{0.75}{
    \begin{tabular}{ccccc}
        \toprule
        \textbf{Model}  & \begin{tabular}[c]{@{}l@{}}\textbf{Activated}\\ \textbf{Experts}\end{tabular} & \begin{tabular}[c]{@{}l@{}}\textbf{Total}\\ \textbf{Experts}\end{tabular} & \begin{tabular}[c]{@{}l@{}}\textbf{Activate}\\ \textbf{Parameters}\end{tabular} & \begin{tabular}[c]{@{}l@{}}\textbf{Total}\\ \textbf{Parameters}\end{tabular} \\
        \midrule
        Phi-MoE  & 2 & 16 & 7.4B & 42B  \\
        OLMoE & 8 & 64 & 1.3B & 6.9B \\
        \bottomrule
    \end{tabular}}
    \caption{Model configurations.}
    \label{tab:models}
    \vspace{-0.4cm}
\end{table}

\textbf{Models.} We evaluate our approach on two latest and representative MoE models: Phi-MoE~\citep{abdin2024phi} and OLMoE~\citep{muennighoff2024olmoe}. These models vary in size and the number of activated experts, as summarized in \Cref{tab:models}.


\begin{table*}[t]
    \centering
    \scalebox{0.75}{
    \begin{tabular}{c|ccccc|cc}
        \toprule
        \begin{tabular}[c]{@{}c@{}}\textbf{Task $(\rightarrow)$} \\ \small{Metric} $(\rightarrow)$\end{tabular} & \begin{tabular}[c]{@{}c@{}}\textbf{IFEval}\\ \small{0-shot Loose Acc}\end{tabular} & \begin{tabular}[c]{@{}c@{}}\textbf{BBH}\\ \small{3-shot EM}\end{tabular} & \begin{tabular}[c]{@{}c@{}}\textbf{TruthufulQA}\\ \small{MC2}\end{tabular} & \begin{tabular}[c]{@{}c@{}}\textbf{GSM8K}\\ \small{0-shot CoT EM}\end{tabular} & \begin{tabular}[c]{@{}c@{}}\textbf{HumanEval} \\ \small{0-shot Pass@1}\end{tabular} & \begin{tabular}[c]{@{}c@{}}\textbf{Avg ($\uparrow$)} \\ \quad \end{tabular} & \begin{tabular}[c]{@{}c@{}}\textbf{Latency ($\downarrow$)} \\ \quad s \end{tabular} \\
        \midrule
        Pretrained & 25.3 & 63.1 & 45.8 & 31.8 & 48.0 & 42.8 & - \\
        Top-2 SFT& 54.9 & 69.5 & 49.3 & 76.7 & 67.1 & 63.5 & 5.59 \\
        \midrule
                Top-1 SFT & 51.4 & 67.1 & 49.2 & 57.6 & 56.9 & 56.4 & 4.01 \\
        Top-1 SFT w/ CE & 53.6 & 67.3 & 48.8 & 65.5 & 64.2 & 60.0 & 4.35 \\\midrule
        Norm. Perf. (\%) & 97.6 & 96.8 & 99.0 & 85.4 & 95.7 & 94.5 & - \\
        \bottomrule
    \end{tabular}}
    \caption{Phi-MoE (pretrained with 2 activated experts) results on general tasks. The inference latency is measured by the time required to process a fixed number of randomly generated tokens in forward passes. Normalized performance measures the relative performance with respect to the full-expert configuration.}
    \label{tab:phi_general}
\end{table*}

\begin{table*}[t]
    \centering
    \scalebox{0.75}{
    \begin{tabular}{c|ccccc|cc}
        \toprule
        \begin{tabular}[c]{@{}c@{}}\textbf{Task $(\rightarrow)$} \\ \small{Metric} $(\rightarrow)$\end{tabular} & \begin{tabular}[c]{@{}c@{}}\textbf{IFEval}\\ \small{0-shot Loose Acc}\end{tabular} & \begin{tabular}[c]{@{}c@{}}\textbf{BBH}\\ \small{3-shot EM}\end{tabular} & \begin{tabular}[c]{@{}c@{}}\textbf{TruthufulQA}\\ \small{MC2}\end{tabular} & \begin{tabular}[c]{@{}c@{}}\textbf{GSM8K}\\ \small{0-shot CoT EM}\end{tabular} & \begin{tabular}[c]{@{}c@{}}\textbf{HumanEval} \\ \small{0-shot Pass@10}\end{tabular} & \begin{tabular}[c]{@{}c@{}}\textbf{Avg ($\uparrow$)} \\ \quad \end{tabular} & \begin{tabular}[c]{@{}c@{}}\textbf{Latency ($\downarrow$)} \\ \quad s \end{tabular} \\
        \midrule
        Pretrained & 16.5 & 32.1 & 35.8 & 12.1 & 18.7 & 23.0 & - \\
        Top-8 SFT & 39.6 & 32.5 & 41.1 & 36.9 & 39.9 & 37.9 & 7.14  \\
        \midrule 
        Top-4 SFT & 34.2 & 30.7 & 39.2 & 33.1 & 36.6 & 34.8 & 5.31 \\        
        Top-4 SFT w/ CE & 35.1 & 31.5 & 41.4 & 35.9 & 38.4 & 36.5 & 5.83 \\
        \midrule
        Norm. Perf. (\%) & 88.6 & 96.9 & 100.7 & 97.3 & 96.2 & 96.3 & - \\
        \bottomrule
    \end{tabular}}
    \caption{OLMoE (pretrained with 8 activated experts) results on general tasks.}
    \vspace{-0.3cm}
    \label{tab:olmoe_general}
\end{table*}

\begin{table}[t!]
    \centering
    \scalebox{0.8}{
    \begin{tabular}{ccc}
        \toprule
        \begin{tabular}[c]{@{}c@{}}\textbf{Task $(\rightarrow)$} \\ \small{Metric} $(\rightarrow)$\end{tabular} & \begin{tabular}[c]{@{}c@{}}\textbf{GSM8K}\\ \small{0-shot CoT EM}\end{tabular} & \begin{tabular}[c]{@{}c@{}}\textbf{HumanEval} \\ \small{0-shot Pass@1}\end{tabular}  \\
        \midrule
        Pretrained & 31.8 & 48.0  \\
        Top-2 SFT & 62.3 & 67.0  \\
        \midrule
        Top-1 SFT & 48.8 & 60.1  \\
        Top-1 SFT w/ CE & 57.4 & 63.1  \\
        \midrule
        Norm. Perf. (\%) & 92.1 & 94.2 \\
        \bottomrule
    \end{tabular}}
    \caption{Phi-MoE results on specialized tasks.}
    \label{tab:phi_special}
    \vspace{-0.3cm}
\end{table}

\begin{table}[t!]
    \centering
    \scalebox{0.8}{
    \begin{tabular}{ccc}
        \toprule
        \begin{tabular}[c]{@{}c@{}}\textbf{Task $(\rightarrow)$} \\ \small{Metric} $(\rightarrow)$\end{tabular} & \begin{tabular}[c]{@{}c@{}}\textbf{GSM8K}\\ \small{0-shot CoT EM}\end{tabular} & \begin{tabular}[c]{@{}c@{}}\textbf{HumanEval} \\ \small{0-shot Pass@10}\end{tabular}  \\
        \midrule
        Pretrained & 12.1  & 14.6  \\
        Top-8 SFT & 32.3 & 39.3  \\
        \midrule
        Top-4 SFT  & 26.8 & 34.7  \\
        Top-4 SFT w/ CE & 29.7 &  38.4 \\
        \midrule
        Norm. Perf. (\%) & 92.0 & 97.7 \\
        \bottomrule
    \end{tabular}}
    \caption{OLMoE results on specialized tasks.}
    \vspace{-0.5cm}
    \label{tab:olmoe_special}
\end{table}

\textbf{Training datasets.} To evaluate compressed experts in various domains and applications, we conduct experiments on three tasks with different scopes. For specialized tasks including mathematics and coding, we finetune models on MathInstruct~\citep{yuemammoth} and Magicoder~\citep{wei2023magicoder}. For general capabilities, we finetune models on T\"{U}LU 3~\citep{lambert2024t}, a general-purpose instruction-following dataset targeting skills including reasoning, math, coding and safety.\looseness=-1

\textbf{Evaluation datasets.} For evaluation, we benchmark models on both specialized and general tasks. For mathematical reasoning and coding, we evaluate on GSM8K~\citep{cobbe2021training} and HumanEval~\citep{chen2021evaluating} respectively. To assess general-purpose capabilities, we further include IFEval~\citep{zhou2023instruction}, TruthfulQA~\citep{lin2021truthfulqa}, BBH~\citep{suzgun2023challenging}. Note that for HumanEval, we use Pass@1 for Phi-MoE and Pass@10 for OLMoE respectively. Since OLMoE is a smaller model with inherently lower capacity for complex coding tasks, Pass@10 provides finer-grained resolution to better capture performance differences across various configurations. The detailed evaluation metric for each task are presented in the tables shown below.

\textbf{Baselines.} Following the methodology outlined in \Cref{sec:emp_val}, we replace half of the activated experts with compressed experts and compare their performance against two baselines: the pretrained model and the configuration with half of the activated experts. Additionally, we report the performance of the full-expert configuration as a performance oracle for reference.

\textbf{Inference latency evaluation.} To evaluate the reduction of inference time with compressed experts, we follow the practice in the latency benchmark presented in \citet{kwon2023efficient}. We create dummy prompts of batch size 8 and sequence length 32 with randomly generated tokens. Then, we pass the dummy prompts to the model and let it generate completions. The model first goes through 10 warmup iterations, and the reported latency is averaged over 30 iterations of completions. \looseness=-1

\subsection{Main Results} \label{sec:results}


We present the performance of compressed experts on general tasks and specialized tasks respectively. \Cref{tab:phi_general} and \Cref{tab:olmoe_general} present the performance on a diverse set of general evaluation tasks, along with the inference latency for each configuration. \Cref{tab:phi_special} and \Cref{tab:olmoe_special} present the performance of Phi-MoE and OLMoE on specialized tasks, including mathematic and coding. Beyond raw performance, we additionally report normalized performance: $\text{perf}_\text{CE}/\text{perf}_\text{full}$. It quantifies how well compressed experts approximate the performance of the full-expert configuration, providing a direct measure of performance retention after expert reduction. 


\textbf{Performance comparison.} The results demonstrate that compressed experts significantly narrow the performance gap with fully activated expert configurations. Across all models and tasks, compressed experts consistently recover more than 90\% of the full expert performance, showcasing their ability to retain key information from auxiliary experts. Additionally, the incorporation of compressed experts consistently outperform the halved baseline. This shows that information contained in auxiliary experts still has a meaningful contribution to the model performance. This allows MoE models to maintain strong task performance with fewer active parameters. Similar to the findings in \citet{huang2024harder}, we observe that the optimal number of activated experts varies based on task complexity and broadness, and compressed experts exhibit task-dependent efficacy, which we detail in \Cref{sec:analysis}.\looseness=-1

\textbf{Inference latency.} Augmenting halved configurations with compressed experts (CE) incurs only marginal overhead. For Phi-MoE, top-1 with CE increases latency by 8.5\% over top-1, while OLMoE’s top-4 with CE adds 9.7\% latency compared to vanilla top-4. This overhead arises from lightweight operations such as element-wise multiplications, which is orders of magnitude cheaper than executing a full expert. Critically, CE-augmented configurations remain far more efficient than their original counterparts: Phi-MoE top-1 with CE is 22\% faster than top-2, and OLMoE top-4 with CE is 18.4\% faster than top-8. Combined with their strong empirical performance, we comprehensively validate that the compressed experts effectively balance efficiency and performance.

\subsection{Empirical Analysis} \label{sec:analysis}

We provide a more detailed analysis on per-task performance in the general evaluation setting (\Cref{tab:phi_general,tab:olmoe_general}). Unlike specialized evaluation with a focus on specific domains, this setting covers factual correctness, logical reasoning, and precise instruction-following abilities. The diversity of tasks makes it challenging for MoE models to optimize for every task simultaneously, which in turn affects how well compressed experts perform. We observe that not all tasks benefit equally from compressed experts. Here, we mainly base our analysis on the performance improvement from the halved baseline.\looseness=-1

\textbf{Compressed experts excel in tasks requiring specialized reasoning.} On mathematical problem-solving (GSM8K) and coding (HumanEval), compressed experts achieve substantial performance gains over configurations with halved experts. These tasks require structured reasoning and pattern-based problem-solving, where compressed experts can approximate the auxiliary experts' role effectively. Since these domains follow well-defined rules (e.g., arithmetic operations, program syntax), the lightweight compressed experts capture task-relevant transformations efficiently, reducing the need for additional full experts.

\textbf{Gains are less pronounced in tasks emphasizing factual recall and instruction following.} In factuality (TruthfulQA) and instruction-following tasks (IFEval), compressed experts offer smaller improvements. This suggests that while compressed experts effectively distill expert knowledge, their lightweight parameterization may not fully capture the extensive factual knowledge distributed across different full experts. As a result, tasks requiring retrieval of fine-grained information might benefit less from expert compression compared to those relying on structured reasoning.

\textbf{Broad reasoning tasks show minimal improvement.} On BBH, which demands diverse knowledge integration and multi-step logical reasoning, compressed experts exhibit similar performance to the baselines with halved expert count on both models. However, this limitation is not specific to compression alone, as the fully activated expert configurations also struggle with these tasks, suggesting that expert count is the primary limiting factor. As noted in \citet{huang2024harder}, such tasks likely require activating a greater number of full experts to synthesize heterogeneous knowledge, which remains constrained in models with fewer active experts. \looseness=-1

Overall, compressed experts provide an effective balance between efficiency and performance, consistently recovering over 90\% of the full expert performance across both models. The performance recovery could vary across different task types, with structured reasoning tasks benefiting more than knowledge-intensive or multi-step reasoning tasks. \looseness=-1

\section{Related Works}
\textbf{Mixture-of-Experts (MoE).} Mixture-of-Experts (MoE) models have emerged as a powerful approach to scaling up model capacity while maintaining computational efficiency. By activating only a subset of experts during inference, MoE models significantly reduce computational overhead compared to dense models. This advantage has led to a surge in open-source MoE implementations, each varying in the number of activated experts per layer. For instance,  Mixtral~\citep{jiang2024mixtral}, JetMoE~\citep{shen2024jetmoe}, OpenMoE~\citep{xue2024openmoe} and Phi-MoE~\citep{abdin2024phi} activate two experts, Qwen-MoE~\citep{yang2024qwen2technicalreport} activates four, DeepSeekMoE~\citep{dai2024deepseekmoe} activates six and OLMoE~\citep{muennighoff2024olmoe} activates eight. However, recent work by \citet{huang2024harder} demonstrates that fully utilizing all activated experts can introduce redundancy for certain tasks, as not all tasks require the full capacity of the top-$k$ experts. Inspired by this insight, we propose compressed experts, designed to compress auxiliary experts during model editing. Compressed experts are model-agnostic and can be seamlessly integrated into existing MoE architectures, including those mentioned above. By reducing the number of activated experts while maintaining performance, our approach enhances computational efficiency without compromising model quality.

\textbf{Soft merging of experts in MoE.} Inspired by the surge of model merging techniques~\citep{ilharco2023editing,wortsman2022model,yadav2023tiesmerging,yang2023adamerging,he2025localizeandstitch}, recent works propose to merge activated experts for more precise gradient computation and faster inference. However, these approaches suffer from a common limitation: the merging operations introduce significant computational overhead, making them impractical for token-based routing. Specifically, those approaches require merging experts on-the-fly based on the per-token routing weights. For instance, SMEAR~\citep{muqeeth2024soft} and Lory~\citep{zhong2024lory} merge experts via averaging the expert parameters based on the routing weights. While effective in sequence-based or semantic-based routing scenarios, the computational cost of these operations renders them unsuitable for token-level routing, where efficiency is critical. In addition, MC-SMoE~\citep{li2023merge} provides a static merged expert configuration shared across all inputs. While this reduces computational costs, it sacrifices the token-level routing flexibility that makes MoE models powerful. In contrast, our method introduces compressed experts, which directly augment hidden state activations. This eliminates the need for dynamic parameter merging. Our approach retains compatibility with existing MoE architectures while offering greater flexibility, as compressed experts avoid the latency penalties of prior merging strategies. \looseness=-1

\textbf{Model pruning.} We highlight two key advantages of our approach over pruning~\citep{lu2024not,he2024demystifying,xie2024moe}, which prunes the MoE model at an expert level. \textbf{i) Higher efficiency}: Pruning is a post-processing method after full training. In contrast, our method integrates compressed experts during training, allowing them to learn from both training data and auxiliary experts. Additionally, besides saving inference, the lightweight structure enables our method to achieve an approximate 20\% reduction in training time as well by eliminating redundant expert computations, whereas post-processing methods like pruning require standard training plus an additional compression step. \textbf{ii) Better knowledge preservation}: Pruning removes experts deemed less important, which can lead to the loss of potentially valuable knowledge, even from experts that contribute less overall (\Cref{sec:km_ka}). In contrast, our goal is to preserve the information, but with smaller capacity.


\textbf{Parameter Efficient Finetuning (PEFT).} With the increasing sizes of LLMs, PEFT~\citep{hetowards} has gained attention due to its capabilities to efficiently tune the model for specific downstream tasks. Instead of tuning the full model, PEFT methods introduce lightweight modules to augment the model, and only update those modules. Popular PEFT approaches include adapters~\citep{houlsby2019parameter}, prompt tuning~\citep{lester2021power}, LoRA~\citep{hulora} and (IA)$^3$\citep{liu2022few}. 

Despite the similarities, compressed expert is \textit{not} a PEFT method. While both approaches leverage lightweight modules, compressed experts focus on reducing computational costs during inference. The compressed experts serve as a compact representation of the auxiliary experts, while the main experts are used in their original form. Specifically, during SFT, the main experts are jointly trained with compressed experts. This design preserves the expressiveness of the main experts while reducing inference costs, addressing a fundamental efficiency-performance trade-off in MoE models.

\section{Conclusions}
We introduce compressed expert, a lightweight and efficient approach to reducing the number of activated experts in MoE models while maintaining strong performance. By replacing auxiliary experts with compact expert representations, our method significantly reduces computational overhead while preserving model capacity. Extensive experiments on Phi-MoE and OLMoE across various tasks demonstrate that compressed experts reduce active parameters by over 30\% and cut inference costs by 20\% while retaining over 90\% of the performance of full-expert configurations. Beyond reducing inference costs, our findings suggest broader implications for scaling MoE models efficiently. Compressed experts offer a promising direction for optimizing sparse activation in MoE architectures. \looseness=-1

\newpage
\section*{Limitations}
Our approach uses a fixed compression ratio, which treats half of the activated experts as main experts, while the other half are auxiliary experts. As shown in \Cref{sec:results}, this configuration may not be optimal for all tasks, especially for models with more than 2 activated experts, which enables more possibilities of main and auxiliary expert distribution. While halving provides a strong efficiency-performance trade-off from our empirical validation in \Cref{sec:km_ka}, some tasks may require a more adaptive compression strategy. To tackle this challenge, future works can explore dynamically determining the degree of expert reduction based on task complexity or input characteristics, and address the additional computational overhead incurred by those approaches.

\bibliography{custom}

\appendix

\section{Active Parameter Calculation} \label{appendix:params}
Here, we provide a similar calculation of active parameter reduction for OLMoE as in \Cref{sec:active_params}. In OLMoE, each MoE layer contains 3 weight matrices of size $1024\times2048$. The total non-MoE parameters are approximately 475M. In the original configuration with 8 activated experts across 16 MoE layers, the number of parameters in MoE layers is
\begin{align*}
    \text{\# MoE params}&=3\times1024\times2048\times16\times8\\
    &\approx 805M.
\end{align*}
With compressed experts, we only activate 4 full experts per layer, and the resulting total MoE parameters is
\begin{align*}
    \text{\# MoE w/ CE params}=&3\times(1024\times2048\times4+ \\
    &\underline{2\times2048+1024})\times16\\
    \approx & 403M.
\end{align*}
In total, the number of active parameters for the original configuration is 1.28B, while the one with compressed experts is only 878M, resulting in a saving of 31.4\%.

\section{Performance-Latency Trade-off} \label{appendix:perf_time}
\begin{figure}[h!]
    \centering
    \includegraphics[width=0.7\linewidth]{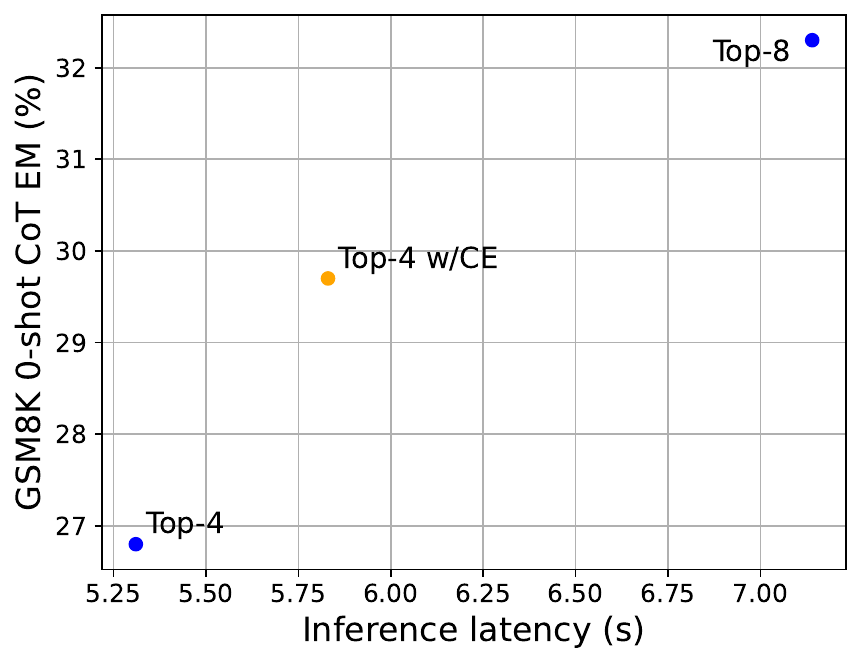}
    \caption{The performance of OLMoE versus the inference latency, each point representing a different expert configuration. The Top-4 w/ CE configuration performs closely to the Top-8 configuration while achieving low inference latency close to Top-4.}
    \label{fig:perf_time_olmoe}
\end{figure}

We produce a similar plot as \Cref{fig:perf_time} for OLMoE. Similar to our observations with Phi-MoE, incorporating compressed experts into the Top-4 configuration strikes a favorable balance between efficiency and performance. Specifically, It only adds a minimal overhead on the inference latency compared with top-4, but noticeably closes the performance gap to top-8. This consistent trend across different models further validates the effectiveness of our approach.

\section{Alternative Compressed Experts Construction}
During the development of our compressed expert design, we have explored alternatives such as autoencoder reductions or LoRA. However, we find that those approaches do not noticeably decrease the inference time. While these methods reduce the parameter count, they still require the execution of full forward passes through MLPs layers, similar to the full experts, thereby retaining much of the original computational overhead. For instance, we compare the inference latency of a LoRA-based implementation with rank 16 for Phi-MoE against our compressed expert approach in \Cref{tab:alter_time}. As shown, LoRA still introduces a noticeable increase in inference time.

\begin{table}[t!]
    \centering
    \scalebox{0.65}{
    \begin{tabular}{lcccc}
        \toprule
        & \textbf{Top-1} & \textbf{Top-1 W/CE} & \textbf{Top-1-W/LoRA} & \textbf{Top-2} \\
        \midrule
        Inference latency (s) & 4.01 & 4.35 & 5.42 & 5.59 \\
        \bottomrule
    \end{tabular}}
    \caption{Inference latency for alternative CE construction.}
    \label{tab:alter_time}
\end{table}

In contrast, our element-wise multiplication approach requires only a simple, low-cost operation to incorporate compressed experts into the model. Empirically, we found that this approach retains sufficient expressive power to capture the critical information from auxiliary experts while substantially lowering inference costs. We will include this discussion in our final manuscript.

\section{Training details}
All experiments are conducted on NVIDIA A100 GPUs. Both models are optimized using the AdamW optimizer~\citep{loshchilov2017fixing} with a cosine learning rate scheduler. To accommodate differences in model scale, the initial learning rate for Phi-MoE is set to 1e-5, while for OLMoE, it is set to 2e-5. The sequence length is fixed at 4096, and the global batch size is 128.

\section{Dataset Details}
The T\"{U}LU 3 dataset is under the ODC-BY-1.0 license. The MathInstruct dataset is under MIT license. The Magicoder dataset is under Apache-2.0 license.

The data does not contain information that can be used to uniquely identify individual people or offensive content.

\section{Potential Risks}
This paper presents work whose goal is to advance the field of NLP. There are many potential societal consequences of our work, none which we feel must be specifically highlighted here.

\end{document}